\def\mytitle{Naive Bayes Entrapment Detection for Planetary Rovers}

\def\myauthor{Dicong Qiu\footnote{\url{http://www.davidqiu.com/}}}
\def\contact{dq@cs.cmu.edu}
\def\Organization{Carnegie Mellon University}

\documentclass[10pt, a4paper]{article}
\usepackage[a4paper,outer=1.5cm,inner=1.5cm,top=1.75cm,bottom=1.5cm]{geometry}
\twocolumn
\usepackage{graphicx}
\graphicspath{{./figures/}}
\usepackage{subcaption}
\usepackage[colorlinks,linkcolor={black},citecolor={blue!80!black},urlcolor={blue!80!black}]{hyperref}
\usepackage[parfill]{parskip}
\usepackage{lmodern}

\usepackage{watermark}
\usepackage{lipsum}
\usepackage{xcolor}
\usepackage{listings}
\usepackage{float}
\usepackage{titlesec}
\usepackage{bm}
\usepackage{hw}
\usepackage{amsmath}
\usepackage{algorithm2e}

\usepackage{enumerate}
\usepackage{enumitem}

\titlespacing{\subsection}{0pt}{\parskip}{-3pt}
\titlespacing{\subsubsection}{0pt}{\parskip}{-\parskip}
\titlespacing{\paragraph}{0pt}{\parskip}{\parskip}

\newcommand{\norm}[1]{\left\lVert#1\right\rVert}

\title{\mytitle}
\author{\myauthor\hspace{1em}\\\contact\\\Organization}
\date{}
\begin{document}

\maketitle

\begin{abstract}
Entrapment detection is a prerequisite for planetary rovers to perform autonomous rescue procedure. In this study, rover entrapment and approximated entrapment criteria are formally defined. Entrapment detection using Naive Bayes classifiers is proposed and discussed along with results from experiments where the Naive Bayes entrapment detector is applied to AutoKralwer rovers. And final conclusions and further discussions are presented in the final section.
\end{abstract}

%\textbf{Keywords -- }{\mykeywords}

\section{Introduction}\label{sec:introduction}

Entrapment detection is a prerequisite for planetary rovers to perform autonomous rescue procedure, because an entrapped rover needs to detect entrapment so as to broadcast rescue request to other rovers. The Roverside Assistance (Symbiotic Planetary Rovers Rescue) project \footnote{\url{http://mrsdprojects.ri.cmu.edu/2017teami/}} aims at demonstrating the idea that collaborative multiple rovers (robots) system is more robust and reliable in some situations of a planetary exploration mission. As a multiple-agent collaboration application, the primary focus of this project is extricating entrapped rovers autonomously.

The entire process of entrapment and extrication in a fully autonomous planetary exploration mission is a typical use case of this project, which includes the following steps:

\begin{enumerate}[label=(\alph*)]

    \item two rovers collaborate together in a planetary exploration mission;
    
    \item one rover, for example the rover AK1, is entrapped during the mission, where the entrapped rover, AK1, is not capable of extricating itself, and such an entrapment is detected autonomously;
    
    \item the entrapped rover, AK1, broadcasts a SOS signal across the rovers network to request rescue;
    
    \item the other rover, for example the rover AK2, receives the SOS signal, so it suspends its current task and become the rescuer rover;
    
    \item the rescuer rover, AK2, approaches the entrapped rover, AK1, in an autonomous manner;
    
    \item after getting close enough to the entrapped rover, AK1, the rescuer rover, AK2, launches the autonomous rescue procedure to extricate the entrapped rover, AK1;
    
    \item after the entrapped rover, AK1, is extricated, the two rovers, AK1 and AK2, resume their original tasks.
    
\end{enumerate}

Entrapment detection (in an autonomous manner) is vital, because:

\begin{enumerate}[label=(\alph*)]

    \item the desired project scope includes entrapment detection and the entire process need to be done autonomously;
    
    \item in a planetary exploration mission, real-time remote monitoring or operation is impossible due to the data transmission delay and limited transmission rate, so rovers shall be capable of autonomously detecting entrapment so as to reduce the potential risks and damages caused by the unawareness of their actual situations and continuing the originally scheduled procedure.
    
\end{enumerate}

\begin{figure}[H]
    \centering
    \includegraphics[width=0.8\textwidth]{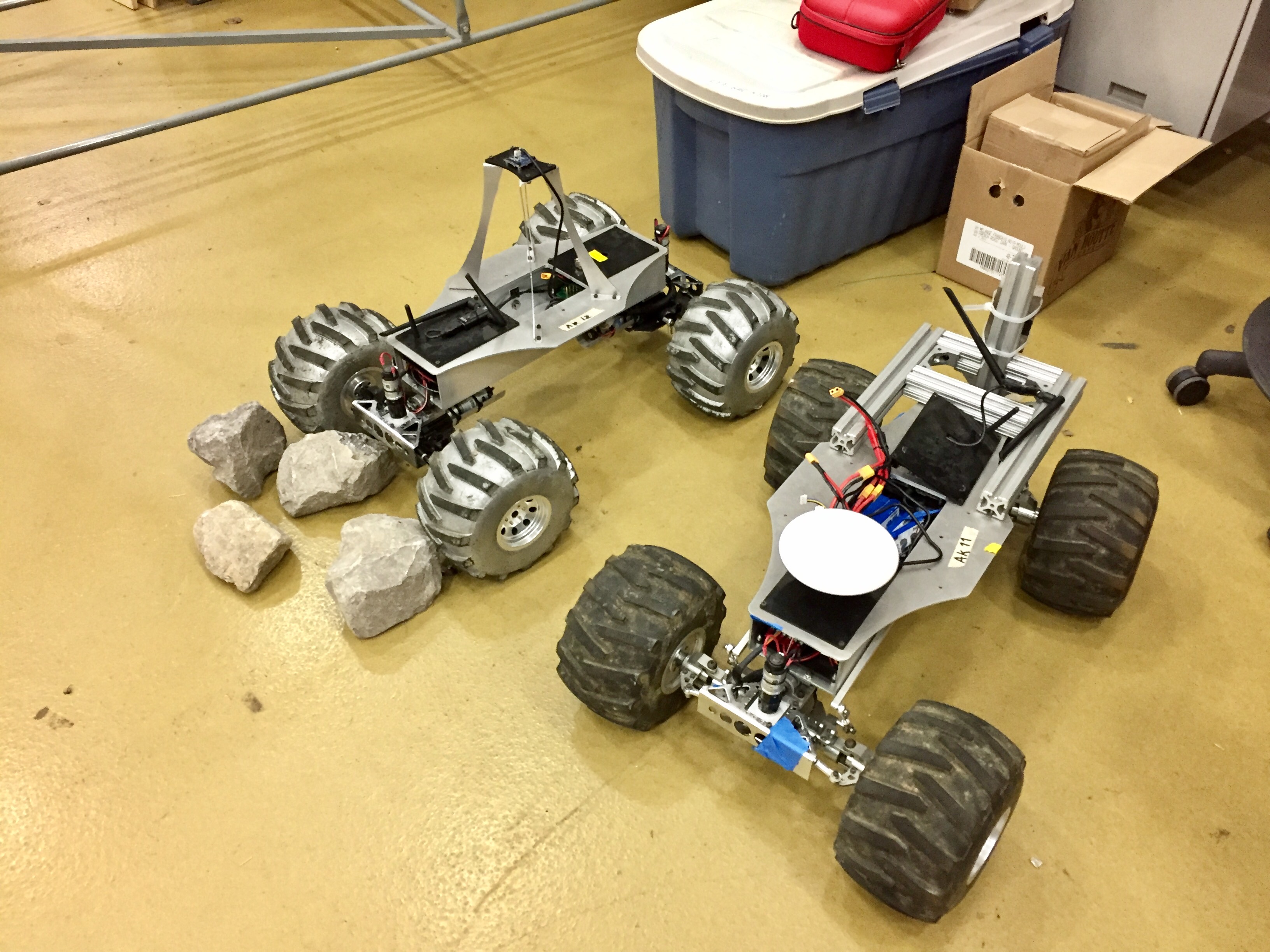}
    \caption{the AutoKrawlers AK1 and AK2}
    \label{fig:autokrawlers}
\end{figure}

In this study of rover entrapment detection, AutoKrawlers, as shown in figure \ref{fig:autokrawlers}, are used as an example of rovers, but the approaches to detect entrapment are generally applicable to different rover models.

In this following sections, the entrapment will be formulated formally, Naive Bayes concept used to estimate the entrapment status of a rover will be discussed, results from real experiments with AutoKrawlers will be shown, and final conclusions and further discussion will be presented in the conclusion section.

\section{Problem Formulation}\label{sec:probformu}

\subsection{Rover Velocities}\label{sec:probformu_velocity}

A rover moves by actuating its locomotion joints, and the state of the locomotion joints can be estimated by joint encoders measurement. By observing the locomotion joint encoder readings, the joint velocities of the rover can be estimated as 

\[
\dot{\mathbf{q}} = [ \dot{q}_{1}, \dot{q}_{2}, \cdots, \dot{q}_{n} ]^{\mathsf{T}}
\]

where $ n $ is the number of locomotion joints (actuators). It is useful to know the relationship between the joint velocities and the rover (task space) velocities.

But before deriving the relationship between the velocity kinematics of the joint space and the task space, it is necessary to analyze the forward kinematics model of the rover, which can be given by the analysis the physical constraints of the rover. For example, the rovers used in the Roverside Assistance project adopts the Ackermann steering control system \cite{mitchell2006analysis}\cite{zhao2013design}, which gives the Ackermann steering constraints. The forward kinematics can be derived by analyzing the relationship between the joint space position and the task space position, and formally expressed as a transformation from joint space variables to task space variables that 

\[
\mathbf{x} = \mathbf{FK}(\mathbf{q})
\]

The velocity kinematics is then derived by applying the chain rule of derivatives 

\[
\dot{\mathbf{x}} = \frac{ \partial \mathbf{ FK }(\mathbf{q}) }{ \partial \mathbf{q} } \dot{\mathbf{q}} = \mathbb{J}(\mathbf{q}) \dot{\mathbf{q}}
\]

where $ \mathbb{J}(\mathbf{q}) $ is the Jacobian of the forward kinematics function with respect to the joint space variable. Since the above task space velocity is derived from the joint space velocity (locomotion actuators velocity), it is not the ground-truth task space velocity of the rover. In this following sections, the notion $ \dot{\mathbf{x}}_{a} $ will be used to refer to the \textbf{assumed rover (task space) velocity} inferred from the locomotion actuator encoder readings, which is also referred to as the \textbf{locomotion actuator odometry}.

In the contrast, the velocity directly observed from sensors that measure the ground-truth motion of the rover is called the \textbf{measured rover (task space) velocity} denoted by $ \dot{\mathbf{x}}_{m} $. And the \textbf{ground-truth rover (task space) velocity} is denoted by $ \dot{\mathbf{x}}_{g} $.

\subsection{Entrapment}\label{sec:probformu_entrapment}

An \textbf{entrapment} of a rover is a situation that the rover actuates its locomotion actuators drives in a specific velocity configuration but the rover does not move. In such a situation, the rover is said to be entrapped in that configuration.

Formally, in a locomotion actuators output configuration $ \dot{\mathbf{q}} $, the rover is entrapped in this velocity configuration if and only if 

\[
\norm{ \dot{\mathbf{x}}_{g} } < 0 + \epsilon_{0} \textit{ and } \norm{ \frac{ \partial \mathbf{FK}(\mathbf{q}) }{ \partial \mathbf{q} } \dot{\mathbf{q}} - \dot{\mathbf{x}}_{g} } > \epsilon_{ag}
\]

where $ \dot{\mathbf{x}}_{g} $ is the ground-truth rover velocity, $ \mathbf{q} $ is the rover locomotion actuator joint position, $ \epsilon_{0} $ and $ \epsilon_{ag} $ are maximum tolerable errors. The first inequality indicates the rover does not move under a tolerable error, and the second inequality indicates the rover velocity estimated from the locomotion actuator odometry diverges from the ground-truth rover velocity beyond a tolerable error. And note that the first term in the second inequality can be rewritten as 

\[
\frac{ \partial \mathbf{FK}(\mathbf{q}) }{ \partial \mathbf{q} } \dot{\mathbf{q}} 
\doteq 
\dot{\mathbf{x}}_{a}
\]

which is the assumed rover velocity derived from locomotion actuator odometry. So the following inequalities are equivalent to those in the entrapment criteria 

\[
\norm{ \dot{\mathbf{x}}_{g} } < 0 + \epsilon_{0} \textit{ and } \norm{ \dot{\mathbf{x}}_{a} - \dot{\mathbf{x}}_{g} } > \epsilon_{ag}
\]

which can also be used as the criteria to define whether a rover is entrapped or not.

The goal of entrapment detection is to detect whether the current rover status matches the above entrapment criteria. This is not an obvious or trivial problem, because it is impossible to access the ground-truth rover velocity $ \dot{\mathbf{x}}_{g} $ and the choice of the maximum tolerable errors $ \epsilon_{0} $ and $ \epsilon_{ag} $ has a great impact on the entrapment detection performance.

\subsection{Entrapment Criteria Approximated with Measured Rover Velocity}\label{sec:probformu_entrapment_measured}

By the definition of entrapment in section \ref{sec:probformu_entrapment}, it is necessary to know the ground-truth velocity of the rover to detect entrapment. However, the ground-truth rover velocity is not accessible from the perspective of a rover, instead, a rover may approximate its ground-truth velocity $ \dot{\mathbf{x}}_{g} $ by the the measured velocity $ \dot{\mathbf{x}}_{m} $ which can be estimated from some sensor(s) so that 

\[
\dot{\mathbf{x}}_{g} 
\approx 
\dot{\mathbf{x}}_{m}
\]

where the error between the ground-truth rover velocity $ \dot{\mathbf{x}}_{g} $ and the measured rover velocity $ \dot{\mathbf{x}}_{m} $ is 

\[
\dot{\mathbf{e}}_{mg} = \dot{\mathbf{x}}_{g} - \dot{\mathbf{x}}_{m}
\]

To reasonably and practically apply this approximation, the magnitude of the above error should be always less than a maximum tolerable error, such that 

\[
\norm{ \dot{\mathbf{e}}_{mg}(t) } = \norm{ \dot{\mathbf{x}}_{m}(t) - \dot{\mathbf{x}}_{g}(t) } < \epsilon_{mg}
\]

where $ \epsilon_{mg} $ is the maximum tolerable error. 

In the case when the above inequality is fulfilled, the ground-truth rover velocity term can be replaced with the measured rover velocity in the original inequalities of the entrapment criteria, and new entrapment criteria are derived, which are formally defined in the following.

Under the condition that 

\[
\norm{ \dot{\mathbf{x}}_{m} - \dot{\mathbf{x}}_{g} } < \epsilon_{mg}
\]

a rover is entrapped in a velocity configuration if and only if 

\[
\norm{ \dot{\mathbf{x}}_{m} } < 0 + \epsilon_{0} 
\textit{ and } 
\norm{ \dot{\mathbf{x}}_{a} - \dot{\mathbf{x}}_{m} } > \epsilon_{ag}
\]

where $ \dot{\mathbf{x}}_{m} $ is the measured rover velocity, $ \dot{\mathbf{x}}_{a} $ is the assumed rover velocity, $ \dot{\mathbf{x}}_{g} $ is the ground-truth rover velocity, and $ \epsilon_{0} $, $ \epsilon_{ag} $ and $ \epsilon_{mg} $ are the maximum tolerable errors. Note that the new criteria above are only applicable if the measured rover velocity is $ \epsilon_{mg} $-close to the ground-truth rover velocity, and the entrapment of a rover has not been defined when this precondition is not fulfilled.

\section{Entrapment Detection with Naive Bayes Classifiers}\label{sec:NB_dector}

Naive Bayes classifiers assume conditional independence between different features and  take advantage of the Bayes rule to estimate posterior probability distributions from prior knowledge. \cite{rish2001empirical}

The basic idea of using Naive Bayes classifiers to detect entrapment is to look at the readings of different sensors and classify the current status of a rover based on these readings. It is assumed that sensor measurements diverge significantly enough in different situations (i.e. rover entrapped, rover slipping, rover stopped, and rover moving normally) so that these situations are probabilistically distinguishable.

\subsection{Velocity Divergence Estimation}\label{sec:NB_detector_divergence}

The velocity divergence status is estimated by the norm of the divergence between the assumed rover velocity $ \dot{\mathbf{x}}_{a} $ and the measured rover velocity $ \dot{\mathbf{x}}_{m} $. In section \ref{sec:probformu_entrapment_measured}, the divergence criterion is approximated by $ \norm{ \dot{\mathbf{x}}_{a} - \dot{\mathbf{x}}_{m} } > \epsilon_{ag} $. To distinguish divergence introduced by linear velocities and angular velocities, linear velocities and angular velocities are separated such that 

\[
\dot{\mathbf{x}}_{a} = \begin{bmatrix}
    \mathbf{v}_{a} \\
    \mathbf{\omega}_{a}
\end{bmatrix}
\textit{ and }
\dot{\mathbf{x}}_{m} = \begin{bmatrix}
    \mathbf{v}_{m} \\
    \mathbf{\omega}_{m}
\end{bmatrix}
\]

where the assumed rover velocity $ \dot{\mathbf{x}}_{a} $ consists of the assumed rover linear velocity $ \mathbf{v}_{a} $ and the assumed rover angular velocity $ \mathbf{\omega}_{a} $, while the measured rover velocity $ \dot{\mathbf{x}}_{m} $ consists of the measured rover linear velocity $ \mathbf{v}_{m} $ and the measured rover angular velocity $ \mathbf{\omega}_{m} $. The approximated velocity divergence criteria is redefined as a weighted error 

\[
Q = \sqrt{
\begin{bmatrix}
    e_{v} & e_{\omega}
\end{bmatrix}
R
\begin{bmatrix}
    e_{v} \\
    e_{\omega}
\end{bmatrix}}
\]

where $ e_{v} = \| \mathbf{v}_{a} - \mathbf{v}_{m} \| $ is the linear velocity error between the assumed linear velocity and the measured linear velocity, $ e_{\omega} = \| \mathbf{\omega}_{a} - \mathbf{\omega}_{m} \| $ is the angular velocity error between assumed and measured angular velocities, and $ R $ is the weight matrix.

The probability whether the assumed rover velocity has diverged from the measured rover velocity 

\[
\mathbf{Pr}(D|Q) = \frac{ \mathbf{Pr}(Q|D) \mathbf{Pr}(D) }{ \sum_{d} \mathbf{Pr}(Q|D=d) \mathbf{Pr}(D=d) }
\]

where $ D \in \{ \textit{diverged}, \textit{consistent} \} $ is the rover velocity divergence status random variable, and $ Q \in \mathbb{R}_{\ge 0} $ is the weighted quadratic divergence between the assumed rover velocity and the measured rover velocity. The prior probability $ \mathbf{Pr}(D) $ is from initialization or the posterior probability $ \mathbf{Pr}(D|Q) $ from the previous estimation.

\subsection{Movement Status Estimation}\label{sec:NB_detector_movement}

The movement status of the rover is estimated by the norm of the measured rover velocity. The probability whether the rover is moving or not 

\[
\mathbf{Pr}(M | \|\mathbf{v}_{m}\|) = \frac{ \mathbf{Pr}(\|\mathbf{v}_{m}\| | M) \mathbf{Pr}(M) }{ \sum_{m} \mathbf{Pr}(\|\mathbf{v}_{m}\| | M=m) \mathbf{Pr}(M=m) }
\]

where $ M \in \{ \textit{moving}, \textit{stopped} \} $ is the rover movement status random variable, and $ \|\mathbf{v}_{m}\| \in \mathbb{R}_{\ge 0} $ is the measured rover velocity norm random variable. The prior probability $ \mathbf{Pr}(M) $ is from initialization or the posterior probability $ \mathbf{Pr}(M | \|\mathbf{v}_{m}\|) $ from the previous estimation.

\subsection{Rover Status Estimation}\label{sec:NB_detector_status}

After the posterior probabilities from sections \ref{sec:NB_detector_divergence} and \ref{sec:NB_detector_movement} are estimated, the probability that the rover is entrapped can be estimated using the conditional independence property of Naive Bayes classifiers.

\begin{equation*}
\begin{split}
      & \mathbf{Pr}(S=\textit{entrapped} | Q, \|\mathbf{v}_{m}\|) \\
    = & \mathbf{Pr}(D=\textit{diverged}, M=\textit{stopped} | Q, \|\mathbf{v}_{m}\|) \\
    = & \mathbf{Pr}(D=\textit{diverged} | Q) \mathbf{Pr}(M=\textit{stopped} | \|\mathbf{v}_{m}\|)
\end{split}
\end{equation*}

where $ S \in \{ \textit{entrapped}, \textit{slipping}, \textit{moving}, \textit{stopped} \} $ is the rover status random variable, and an entrapment is assumed if the joint condition is met that the rover velocity is diverged and the measured rover velocity suggests that the rover is stopped. A bonus of adopting this approach is that, instead of giving only the entrapment status estimation, it provides a complete rover status estimation along with probabilities or confidence levels assigned to each status.

\section{Experiments on AutoKralwers}\label{sec:experiments}

AutoKrawlers are used as the experiment platform to deploy the entrapment detection algorithm and conduct experiments on. AutoKrawlers are equipped with wheel encoders and HTC Vive \cite{niehorster2017accuracy} trackers, where the wheel encoders provide wheel odometry with raw data, which will is then the data source of assumed rover velocity $ \dot{\mathbf{x}}_{a} $, while HTC Vive tracking systems provide another independent measurement of the rover velocity as the measured rover velocity $ \dot{\mathbf{x}}_{m} $.

Instead of performing online learning and testing, the AutoKrawler data sets\footnote{\url{http://mrsdprojects.ri.cmu.edu/2017teami/dataset/}} are collected beforehand, so that the learning and testing conditions are consistent. AutoKrawler data sets are collected with in the following scenarios:

\begin{itemize}
    \item rover entrapped and jiggling, which simulates the scenarios where a rover is partially entrapped, but it is still able to perturb its own pose by some control output;
    \item rover strictly high-centered, which simulates the scenarios where a rover is fully entrapped, and its control output can hardly have significant perturbation on its own pose;
    \item rover moving on a flat surface, which simulates the scenarios where the surfaces to perform locomotion on are relatively flat, and the noise is relatively small;
    \item rover moving on a rocky surface, which simulates the scenarios where a rover needs to perform locomotion on rough surfaces with rocks and gravels, and the noise is relatively large.
\end{itemize}

\subsection{Probabilistic Models}\label{sec:experiments_models}

The probabilistic models (conditional probability models) for computing the velocity divergence status posterior probability in section \ref{sec:NB_detector_divergence} and the rover movement status posterior probability in section \ref{sec:NB_detector_movement} can be learned from data collected from AutoKrawlers in different scenarios.

\begin{figure}[H]
    \centering
    \begin{subfigure}{0.45\textwidth}
        \includegraphics[width=\linewidth]{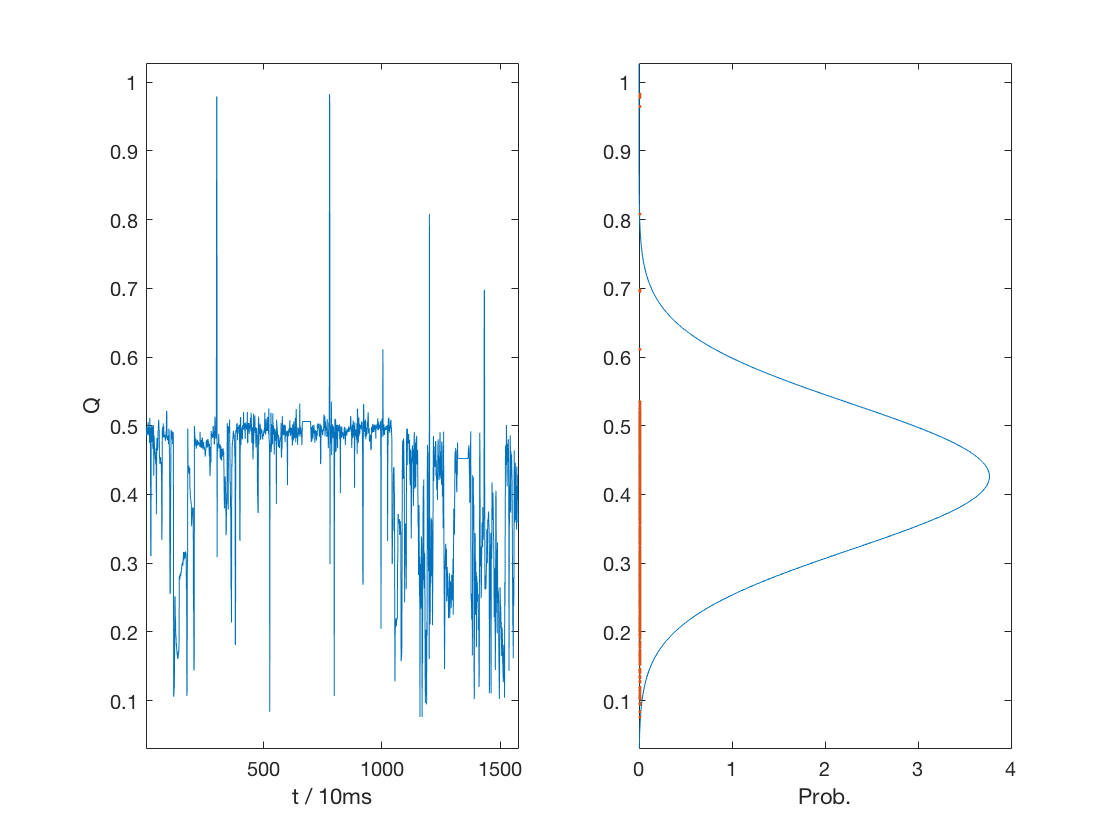}
        \caption{velocity divergence along the time and corresponding probabilistic model fitting result}
        \label{fig:divergence_status_diverged_fitting}
    \end{subfigure}
    \begin{subfigure}{0.45\textwidth}
        \includegraphics[width=\linewidth]{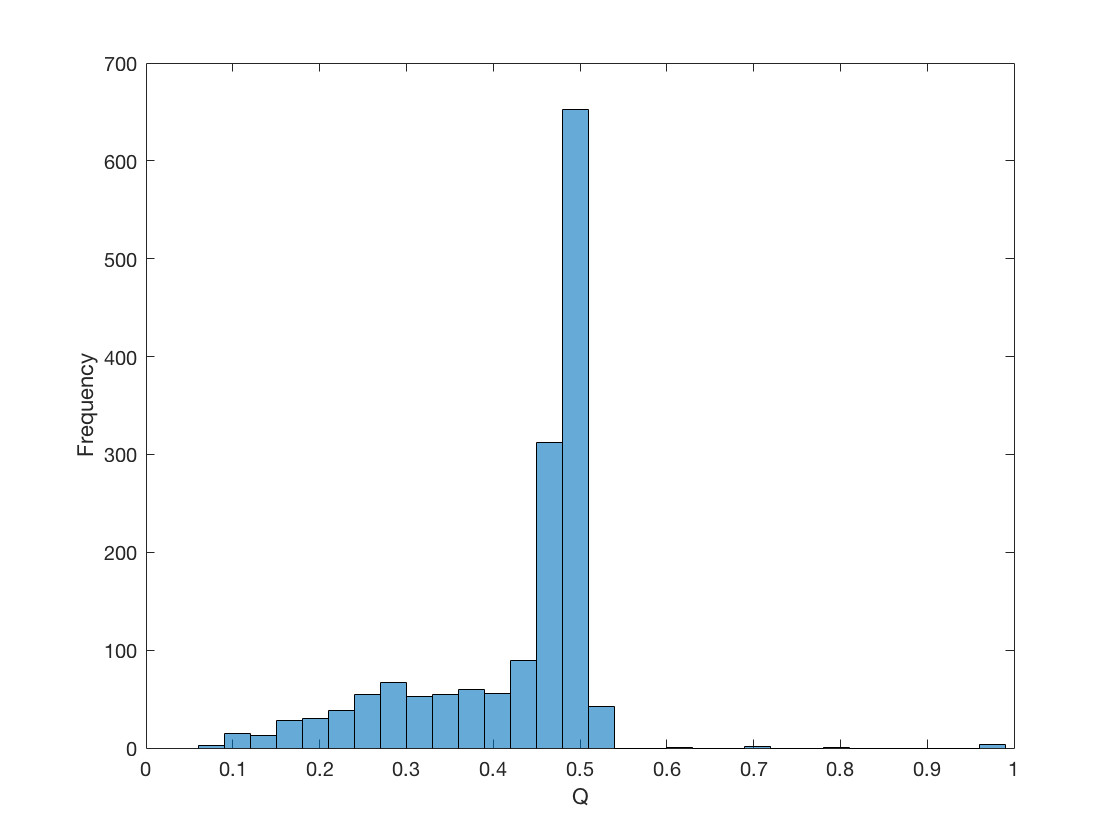}
        \caption{velocity divergence distribution}
        \label{fig:divergence_status_diverged_dist}
    \end{subfigure}
    \caption{diverged velocities scenario for velocity divergence status estimation consisting of both strictly high-centered and stuck-and-jiggling situations, where the low divergence data (with $ Q \le 0.075 $), introduced during the transience of control command switching, are removed during pre-processing}
    \label{fig:divergence_status_diverged}
\end{figure}

\begin{figure}[H]
    \centering
    \begin{subfigure}{0.45\textwidth}
        \includegraphics[width=\linewidth]{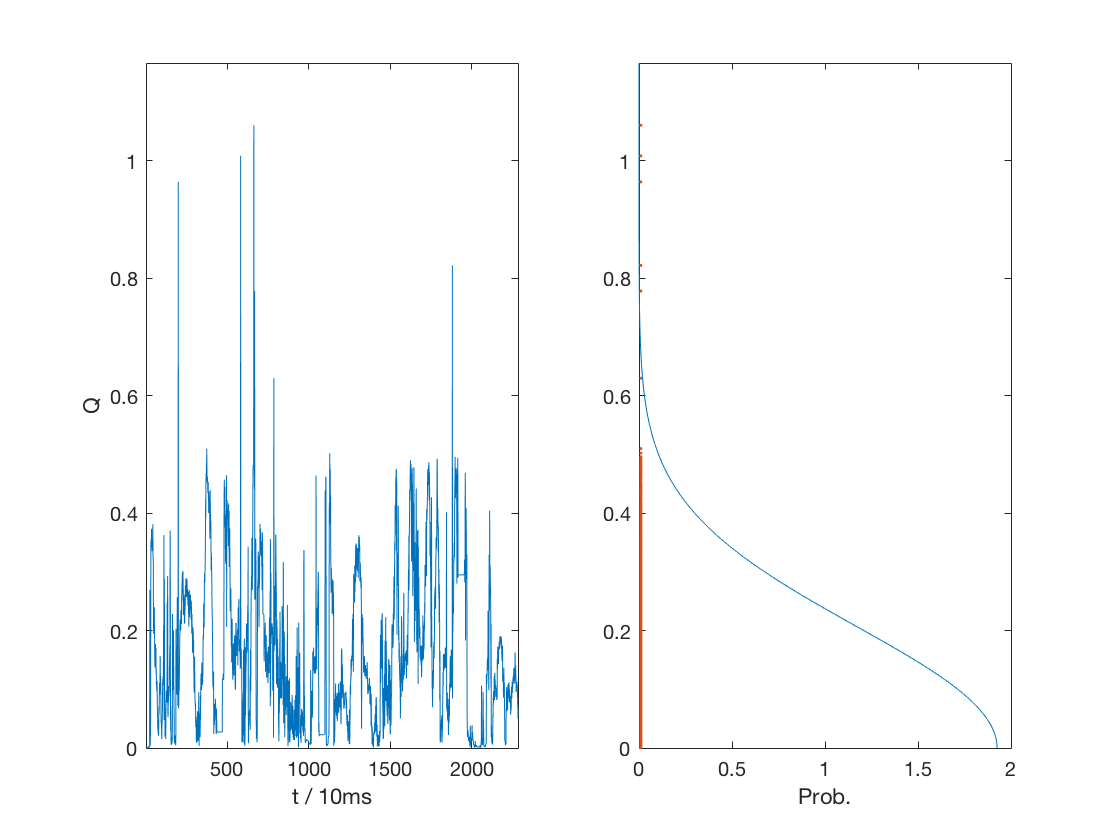}
        \caption{velocity divergence along the time and corresponding probabilistic model fitting result}
        \label{fig:divergence_status_consistent_fitting}
    \end{subfigure}
    \begin{subfigure}{0.45\textwidth}
        \includegraphics[width=\linewidth]{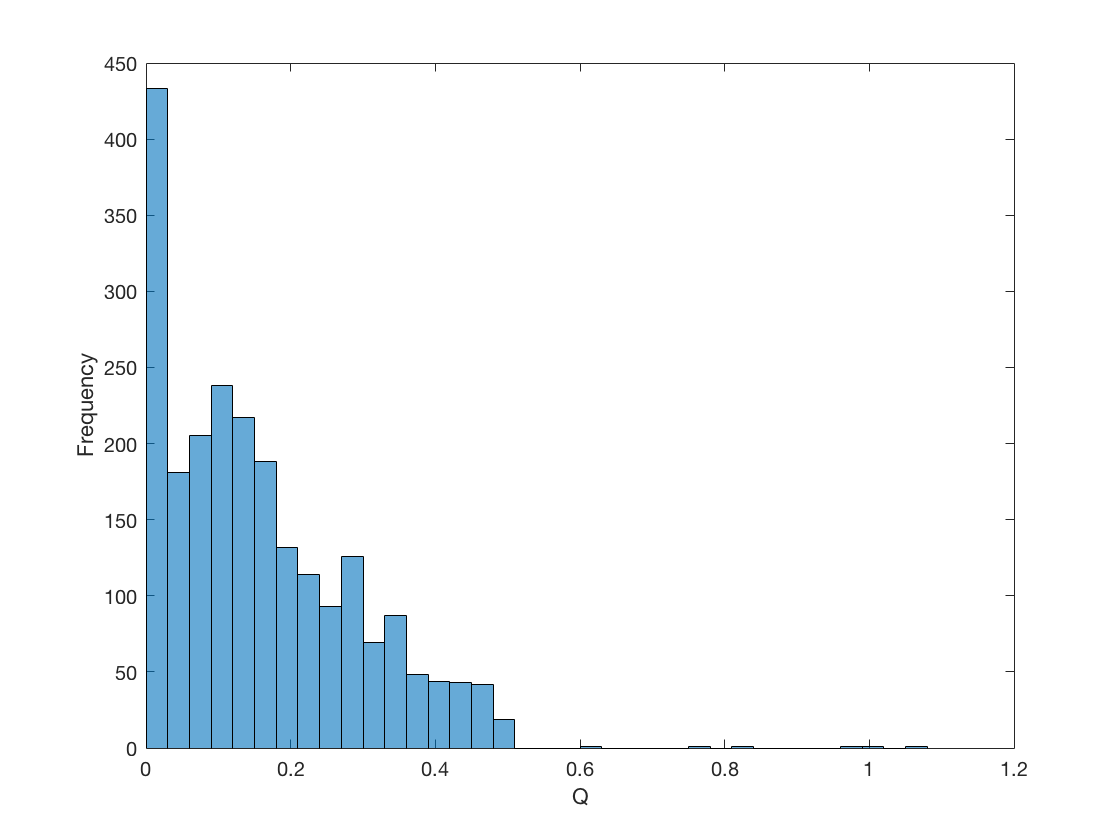}
        \caption{velocity divergence distribution}
        \label{fig:divergence_status_consistent_dist}
    \end{subfigure}
    \caption{consistent velocities scenario for velocity divergence status estimation, which consists of locomotion on both flat and rocky surfaces}
    \label{fig:divergence_status_consistent}
\end{figure}

The conditional probability distribution $ \mathbf{Pr}(Q|D) $ in section \ref{sec:NB_detector_divergence} of weighted error $ Q $ given velocity divergence status $ D $, is assumed to be a Gaussian distribution, with $ \mu = 0.426055 $ and $ \sigma^{2} = 0.011208 $ after data fitting, when $ D = \textit{diverged} $, and is assumed to be a semi-Gaussian distribution, with $ \mu = 0.000000 $ and $ \sigma^{2} = 0.042947 $ after data fitting, when $ D = \textit{consistent} $. The velocity divergence distributions of the data collected from AutoKrawlers and corresponding model fitting results are presented in figures \ref{fig:divergence_status_diverged} and \ref{fig:divergence_status_consistent}, respectively.

\begin{figure}[H]
    \centering
    \begin{subfigure}{0.45\textwidth}
        \includegraphics[width=\linewidth]{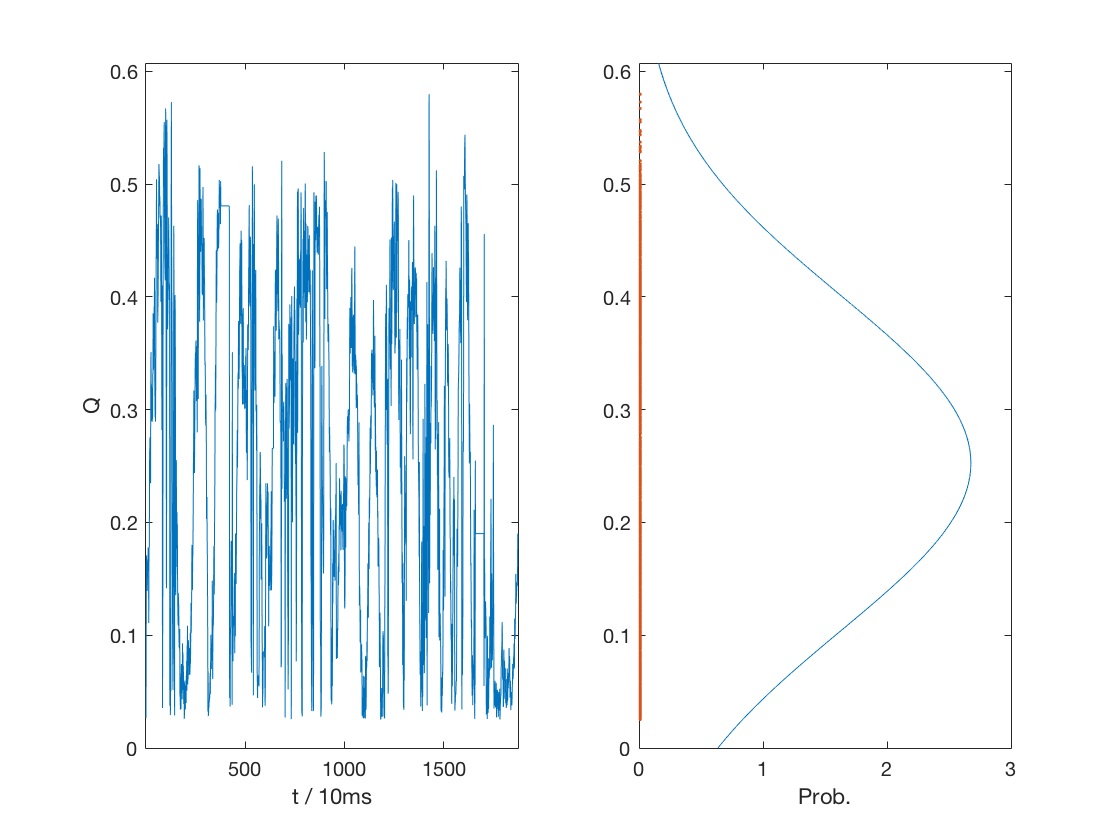}
        \caption{measured linear velocity norm along the time and corresponding probabilistic model fitting result}
        \label{fig:movement_status_moving_fitting}
    \end{subfigure}
    \begin{subfigure}{0.45\textwidth}
        \includegraphics[width=\linewidth]{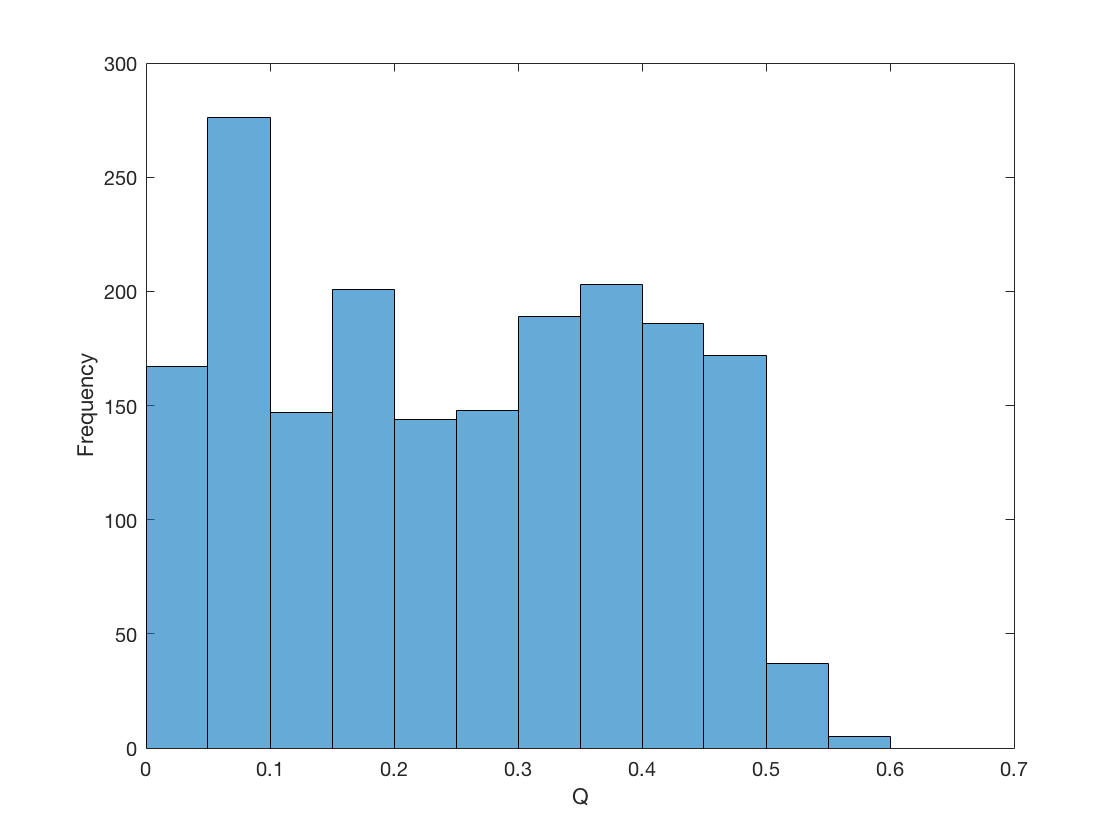}
        \caption{linear velocity norm distribution}
        \label{fig:movement_status_moving_dist}
    \end{subfigure}
    \caption{normally moving scenario for rover movement status estimation, which consist of locomotion on both flat and rocky surfaces, where the measured linear velocity norm is estimated directly from the HTC Vive tracking system}
    \label{fig:movement_status_moving}
\end{figure}

\begin{figure}[H]
    \centering
    \begin{subfigure}{0.45\textwidth}
        \includegraphics[width=\linewidth]{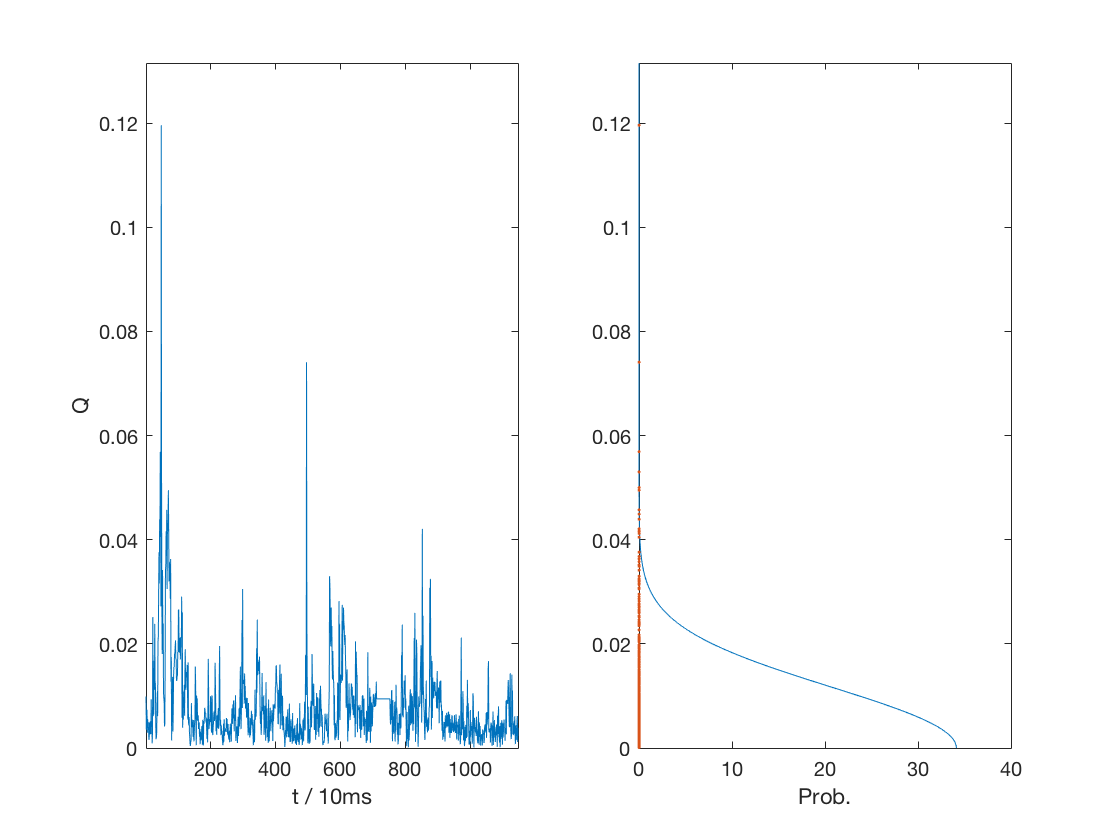}
        \caption{measured linear velocity norm along the time and corresponding probabilistic model fitting result}
        \label{fig:movement_status_stopped_fitting}
    \end{subfigure}
    \begin{subfigure}{0.45\textwidth}
        \includegraphics[width=\linewidth]{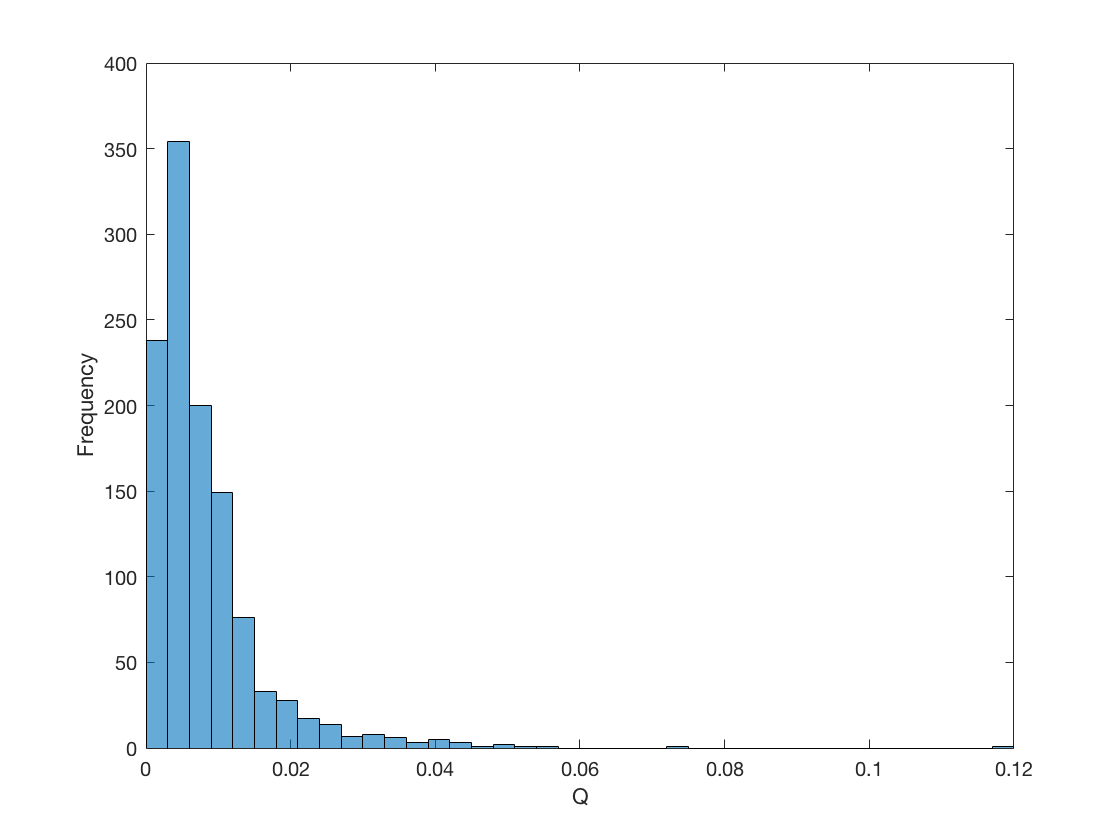}
        \caption{linear velocity norm distribution}
        \label{fig:movement_status_stopped_dist}
    \end{subfigure}
    \caption{rover stopped scenario for rover movement status estimation, where the strictly high-centered situation is used but only the measured linear velocity norm is considered, which is estimated directly from the HTC Vive tracking system}
    \label{fig:movement_status_stopped}
\end{figure}

The conditional probability distribution $ \mathbf{Pr}(\|\mathbf{v}_{m}\| | M) $ in section \ref{sec:NB_detector_movement} of measured rover velocity magnitude $ \|\mathbf{v}_{m}\| $ given rover movement status $ M $, is assumed to be a Gaussian distribution, with $ \mu = 0.252618 $ and $ \sigma^{2} = 0.022222 $ after data fitting, when $ M = \textit{moving} $, and is assumed to be a semi-Gaussian, with $ \mu = 0.000000 $
and $ \sigma^{2} = 0.000137 $ after data fitting, distribution when $ M = \textit{stopped} $. The measured rover velocity magnitude distributions of the data collected from the HTC Vive tracking system and corresponding model fitting results are presented in figures \ref{fig:movement_status_moving} and \ref{fig:movement_status_stopped}, respectively.

\subsection{Entrapment Detection}\label{sec:experiments_test}

\begin{figure}[H]
    \centering
    \includegraphics[width=0.8\linewidth]{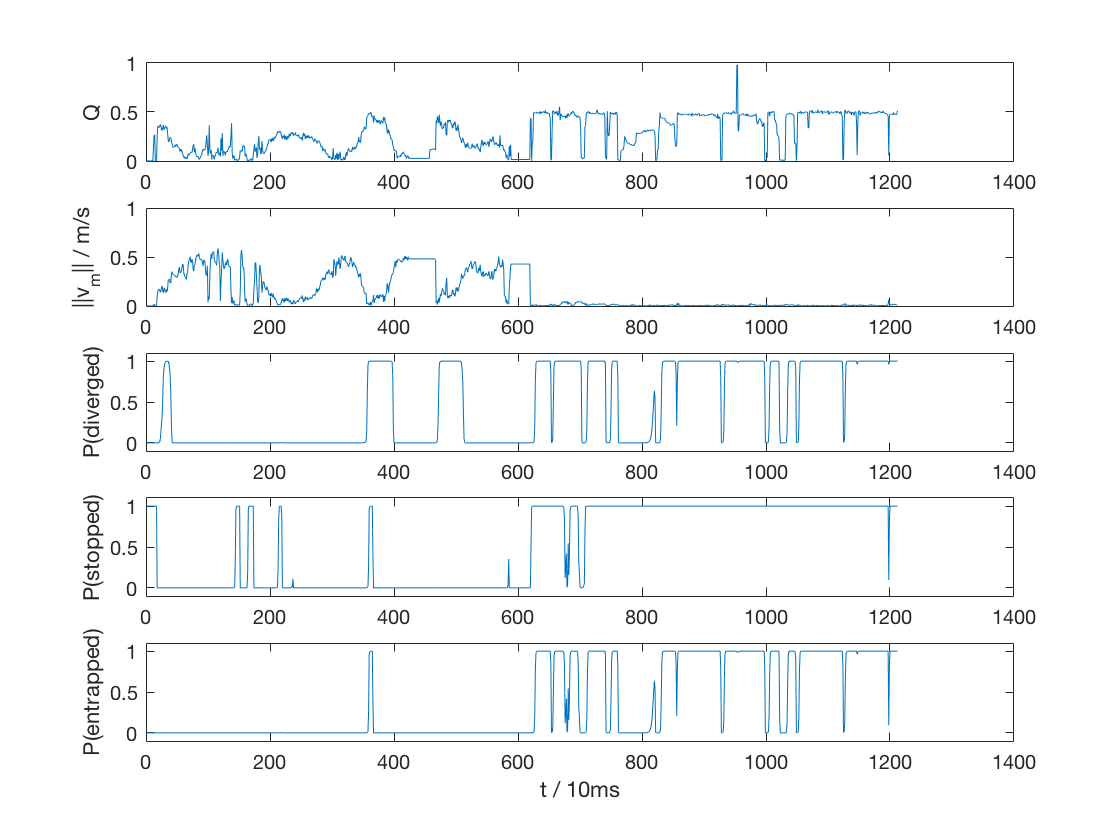}
    \caption{probabilistic entrapment detection result, where the rover was moving normally until entrapped (strictly high-centered) at around $ t=600 $}
    \label{fig:entrapment_detection_prob}
\end{figure}

The entrapment detector is tested with trained models mentioned in section \ref{sec:experiments_models}. In the test data set, the AutoKrawler moved normally on a flat surface for the first half $ 600 \times 10 \textit{ms} $, and got entrapped (strictly high-centered) at around $ t \approx 600 \times 10 \textit{ms} $. The entrapment detector prediction along the time is show in figure \ref{fig:entrapment_detection_prob}, along with intermediate data. %raw weighted error $ Q $, measured rover velocity magnitude $ \| \mathbf{v}_{m} \| $, and the probabilistic predictions of velocity divergence status and movement status.

\section{Conclusion}\label{sec:conclusion}

The Naive Bayes approach conduces to the justification and usability of the entrapment detector by several advantages:

\begin{enumerate}[label=(\alph*)]

    \item Naive Bayes classifiers classify different classes (i.e. whether a rover is entrapped or not) by looking at the corresponding probabilities of different classes and choosing the one with the highest probability as its final prediction. It generates not only the final prediction (entrapment detection), but also the corresponding probability or confidence level.
    
    \item The measurements from different sensors are independent because they come from different sources. This property naturally conform to the "Naive" (conditional independence) assumption of Naive Bayes classifiers.
    
\end{enumerate}

The difficulty lies in the choice of data sources (sensors) and how to properly transform the data (sensor measurements) such that they are applicable to this method. In section \ref{sec:experiments}, data source combination of wheel odometry and the HTC Vive tracking system odometry is adopted. However, in practical and realistic scenarios, such as a planetary exploration mission, it is unfeasible to assume an HTC Vive tracking system pre-installed as an environmental facility. Instead, visual odometry \cite{maimone2007two, corke2004omnidirectional, dinesh2013improvements} can be served as an alternative reference data source for the entrapment detection task.

\section*{Acknowledgement}

This study is made possible thanks to the great support from all members of the Moon Wreckers team\footnote{\url{http://mrsdprojects.ri.cmu.edu/2017teami/team/}}, Eugene Fang, and the Carnegie Mellon University Field Robotics Center. Especially, Prof. William "Red" L. Whittaker, advisor for the Roverside Assistance project, is acknowledged and appreciated for his advice and commons on our project and his encouragement on our team, the Moon Wreckers.

\bibliographystyle{ieeetr}
\bibliography{ms}

\newpage

\onecolumn

\section*{Appendix I: Figures}

\end{document}